------------

**Title:**

# Systematic quantitative analyses reveal the folk-zoological knowledge embedded in folktales

**Authors:**

Yo Nakawake[1,2*], Kosuke Sato[3*]

[1] Institute of Cognitive and Evolutionary Anthropology, University of Oxford

[2] The Graduate School of Human-Environment Studies, Kyushu University

[3] Organization for the Strategic Coordination of Research and Intellectual Properties, Meiji University

[*] Those authors equally contributed. To whom correspondence should be addressed to Yo Nakawake (yo.nakawake@anthro.ox.ac.uk) or Kosuke Sato (sato_kosuke@meiji.ac.jp)

**Abstract**

Cultural learning is a unique human capacity essential for a wide range of adaptations. Researchers have argued that folktales have the pedagogical function of transmitting the essential information for the environment. The most important knowledge for foraging and pastoral society is folk-zoological knowledge, such as the predator-prey relationship among wild animals, or between wild and domesticated animals. Here, we analysed the descriptions of the 382 animal folktales using the natural language processing method and descriptive statistics listed in a worldwide tale-type index (Aarne-Thompson-Uther type index). Our analyses suggested that first, the predator-prey relationship frequently appeared in a co-occurrent animal pair within a folktale (e.g., cat and mouse or wolf and pig), and second, the motif of 'deception', describing the antagonistic behaviour among animals, appeared relatively higher in 'wild and domestic animals' and 'wild animals' than other types. Furthermore, the motif of 'deception' appeared more frequently in pairs, corresponding to the predator-prey relationship. These results corresponded with the hypothesis that the combination of animal characters and what happens in stories represented relationships in the real world. The present study demonstrated that the combination of quantitative methods and qualitative data broaden our understanding of the evolutionary aspects of human cultures.

**Systematic quantitative analyses reveal the folk-zoological knowledge embedded in folktales**

# 1. Introduction

## 1.1 Homo Fabulans

Folktales are cultural universals, widely observed from hunter-gathering society to modern society [1], and popular folktales such as 'Cinderella' or 'Little Red Riding Hood' have been found in cultural groups throughout the world [2,3]. A recent cultural phylogenetic analysis, a statistical method in evolutionary biology, revealed that worldwide folktales can be traced back to 7000 years ago, even before the birth of writing systems [4,5]. Traditionally, folktales were transmitted via verbal communications, but today folktales are transmitted and consumed via a wide range of media such as books, performing arts, and movies, and folktales are enjoyed as a form of entertainment in every type of society [2,3] . In hunter-gatherer societies, people spend considerable time around campfires at night, which provides an opportunity to share folktales [6,7], and anthropologists have argued that folktales contain various forms of knowledge essential for adaptation to both social and natural environments [6, 8–11].

One important role of folktales is to increase sociality [6,7]. While telling folktales has the direct function of social grooming [7], folktales also contain social knowledge such as moral lessons that can enhance group communication. In the Agatha community of the Philippines, folktales contain cooperative values (e.g., equality), which foster coordination and cooperation in foraging societies; furthermore, communities with skilled storytellers had higher rates of cooperation than other societies [11]. These studies suggest that folktales have a pedagogical function to enforce social norms, which increases human adaptability to the local environment [9,10].

The other role of folktales is to increase adaptability to the natural environment. Scalise-Sugiyama has argued that many animal folktales contain folk-zoological knowledge about harmful predators and the nature of animals. Although Scalise-Sugiyama listed several cultural societies that transmit folk-biological knowledge in folktales [12], as far as we know, systematic and quantitative analyses have not yet been conducted on folk-zoological knowledge. In the current study, we hypothesise that folk-biological knowledge is embedded in folktales in such a way that characters and events that appear in folktales represent the real-world environment. Below, we briefly explain how the transmission of folk-zoological knowledge is adaptive.

**1.2. Story and culture transmission**

Knowledge about harmful animals is very important for human adaptation [8,12]. These harmful animals can directly attack humans or human property, such as domestic animals. Some zoological knowledge, including the habitus of animals and predator-prey relationships (e.g., wolves and sheep) appears to be well-known and rather intuitive. However, these intuitive pieces of knowledge are extremely costly if they are learned individually via direct observation or through the experience of being attacked [8,12]. The possibility of identifying predator animals is likely to be limited as identifying the species based on backward induction requires knowledge of the species.

Individual learning requires at least the experience of being attacked or witnessing an attack, and subsequently, social learning reduces the individual learning cost. Social learning, learning from observations made by others or listening to stories/episodes allow knowledge to be transmitted at a low cost [13,14]. This is because social learning does not require the individual experience cost (e.g., an individual being attacked or an individual's domestic animals being attacked). This learning ability from others is unique to humans because of the highly communicative skills that we possess, such as language or reading the intentions of others, allow us to transmit information faithfully [15–17]. Empirical studies have demonstrated that cultural transmission through language allowed knowledge to be transmitted in high fidelity through skill learning compared to other modalities (e.g., direct observation) [18]. Researchers have also argued that folktales are one of the most effective methods of transmitting knowledge effectively through languages [3,8,10,11].

The benefit of human cultural transmission is not only low-cost learning with high fidelity, but there are also cumulative features [15–17]. An individual can learn and produce only limited knowledge by themselves but learning in society allows individuals and societies to acquire knowledge efficiently through copying and teaching, and thus, societies can produce further advanced knowledge, cumulatively compared to previous generations [15,19,20]. This cumulative feature of human culture is known as the 'ratchet effect' (i.e., the instrument to prevent society from going back once it becomes advanced) [15]. The transmission of cultural knowledge such as through tool-use can be found in the animal kingdom, but its cumulative feature only appears in humans. The evidence has demonstrated that storytelling and story reading increase cognitive skills such as language acquisition, understanding other people's thoughts, and sharing beneficial knowledge as pedagogical functions [9]. Storytelling or reading also enhances children's language abilities and understanding other people's thoughts [21,22].

Another advantage of telling folktales is cognitive attraction [10,23]. Ethnographic evidence has demonstrated that folktales are often told in certain rhythms, with a ritualised style, or with redundancy (repetition of words) to attract children [6,10,24,25]. The content of folktales often contains memorable and attractive content. Another form of cognitively attractive content is counterintuitive characters such as fairies or talkative animals, which are known to be highly memorable compared to intuitive characters. Counterintuitive characters are found in the majority of worldwide folktales [26], and popular folktales such as the Brothers Grimm that tend to contain counterintuitive characters, receive more attention on the Internet [27]. Among the counterintuitive characters, the most frequent are counterintuitive animals such as talkative animals [26]. Zipes argued that some of the most well-known current folktales involve characters that may change their shape during the transmission of the story because they are more cognitively attractive to humans [3]. Thus, cognitive attraction or catching people's attention functions as an effective tool to convey cultural knowledge.

### 1.3. Cultural evolution of stories

Folktales function to transmit cultural knowledge, but the folktale itself is the culture that is transmitted and subsequently, it changes shape because of the processes involved in what has been termed cultural evolution. When researching cultural shifts, cultural evolution is used as an analogy of biological evolution by defining culture as the behavioural traits socially transmitted from one individual to another. This approach has been adapted to develop statistical methods to analyse the evolutionary biology of cultures. This means that using the analogy of biological evolution, various theories in different social sciences fields can be empirically tested to investigate cultural shifts, such as the morphology of arrowheads [28] or religious rituals [29].

Folktales are a type of cultural material that has been studied rigorously in cultural evolution studies [19]. The theories of cultural evolution have been tested in general, such as how geography and cultural boundaries influence cultural variations [30], or how population diversity has influenced the variation of folktales [31]. Furthermore, the application of phylogenetic methods on a certain type of folktale made it possible to reveal the origin of the folktale, both for the period and regions [4,5]. This new quantitative approach opens the possibility to infer the source or the period of the folktale objectively.

While this new cultural evolution approach creates new methods to test cultural evolution theories and folkloristics empirically, it does not utilise the rich number of texts that originally interested folklorists. Traditionally folklorists use the method known as 'close reading', which requires experienced researchers to read the text carefully word by word. Computational folkloristics, another approach to study folktales, analyses the texts with the method known as 'distant reading' [32]. Distant reading is the method to apply quantitative and machine-powered analyses (e.g., natural language processing) to the massive amount text, which would be difficult to 'close read' in a finite or limited period.

In this study, to reveal the knowledge embedded in folklore texts, we analysed the Aarne-Thompson-Uther type index (ATU) [33], a collection of international folktale abstracts that have been used in cultural evolution research [4,30] using the computational folkloristic approaches of natural language processing and statistical analyses.

We investigated the folk-biological knowledge embedded in folktales using co-occurrence and motif analyses. First, we conducted a co-occurrence analysis of animals. As we have argued that folktales contain folk-biological knowledge useful to the real world, we hypothesised that the co-occurrence of the animals represents relationships in the real world. Second, we conducted a motif analysis. As we do not know what this co-occurrence of animals signifies, we conducted an additional analysis; the co-occurrence may signify allies, hostile relationships, or neutral relationships. We focused specifically on the motif of 'deception'. This is because this motif represented hostile relationships, which often appear in nature as the predator-prey relationship as they often deceive each other in the animal kingdom.

## 2. Method

### 2.1. Corpus

We used the Aarne-Thompson-Uther type index (ATU) as our corpus, the set of text data to be analysed using natural language processing. The ATU is a catalogue of world folktales, and they are categorised by the 'tale type', the plot pattern of tale. It was originally organized by the Finnish folklorist Antti Aarne, and further revisions were made by Stith Thompson and then Hans-Jörg Uther (ATU is the abbreviation of these names of editors). The feature of the ATU is that ATU numbers are assigned for each tale type; for example, the tale type 'fox and crane invite each other' were assigned ATU60, and this is referred to as its index. In this research, we used Uther (2004), which is the latest English version of the ATU [33]. We used the tale types labelled ATU 1 - ATU 299, which are classified as 'animal tales'. These animal tales consist of five subcategories within the main category of animals (e.g., wild, domestic; see Table 2.1).

Table 1. Index numbers and minor categories in animal tales (ATU 1-299)

| Index | Minor Category | *N* of tale types |
|---|---|---|
| 1 - 99 | Wild Animals | 110 |
| 100 -149 | Wild Animals and Domestic Animals | 72 |
| 150 - 199 | Wild Animals and Humans | 59 |
| 200 - 219 | Domestic Animals | 35 |
| 220 -299 | Other Animals and Objects | 106 |

The corpus contains 462 tale types. The number of indices and the total count of tale types do not match because some indices such as ATU111 have multiple variants such as ATU111a/ATU111b. There are also indices that are allocated to certain numbers but are yet to be allocated to certain types of folktale. As we aimed to analyse the description of the folktales, we excluded tale types that had no description of tale content (e.g., they simply referred to another tale type such as 'see ATU XX'), thus 382 tale types with descriptions made up the final corpus to be analysed. To create machine-readable datasets of the corpus, we digitised a physical copy of the Uther (2004) by using OCR (Optical Character Recognition). The scanned data were then checked and corrected manually.

### 2.2. Structure of the corpus

The structure of each tale type consists of (1) the title and content, (2) additional information, and (3) the bibliography and regional information:

(1) The title and content include the name of the tale type and a summary description of the story. This description is annotated with motifs for each time the texts of a description are matched to a certain motif. A motif is a repeated story element, such as a certain character, a concept, or an event that occurs in stories frequently [34]. For example, motifs can be objects such as the devil or angel, and they can be an action or event such as the 'Creation of the Earth'. Each motif was tagged with the Thompson Motif Index (TMI), a systematic classification developed by Thompson (1955) [34]. TMI classified motifs into major categories by alphabetical letter from A to Z, and each subordinate item was represented by a number (e.g. A.1) For example, the letter 'A' stands for motifs related to 'mythical motifs' and TMI A2200–A2599 is the category group that represents the motifs that are 'an explanation of the origins on animals'. While some tale types have more than one motif, there are some tales that have no motifs.

(2) Additional information includes the editors' notes on whether the story (tale type) is part of famous collections such as 'Aesop's Fables' and also if the tale type is part of a combination with other tale types.

(3) Bibliography and regional information include where the folklore was recorded and also its bibliography.

### 2.3 Data Processing

We extracted animal and motif information from the corpus. To extract data from the raw text file, we used Python (ver. 3.6.3) and the NLTK (Natural Language ToolKit [35]) module (ver. 3.2.4).

#### 2.3.1 The occurrence of animals

To count animal occurrence, we processed the corpus in three steps. First, the animals were extracted from the corpus and listed. We extracted all of the words in the corpus, and then extracted the nouns classified as ‘animal’. For classification, we used WordNet (ver.3.0) data included in the NLTK module. Wordnet is a lexical database of the English language and records the semantic relationships of words [36]. We collected words that contain ‘animal’ in their hypernym, as the superordinate concept. For example, the word ‘dog’ is within the category of ‘animal’, and thus ‘dog’ is a subordinate concept, and ‘animal’ is the superordinate concept. If a word was semantically related to animals as described in WordNet, it was a hypernym of ‘animal’. These extracted words were verified manually, and if it was not an animal’s name, they were excluded.

Second, the list was verified by unifying the same animals but using different words and then coding these animal names. Animals that had various names were unified to a certain name (e.g., cock, rooster, and hen as ‘chicken’), and some animal categories that appeared less frequently (n < 5) were unified (e.g., thrash, woodpecker, and parrot as ‘bird’).

Finally, we counted the animals using the verified list. In this procedure, some verbs that could be confused with an animal name, such as ‘fly’, were manually checked and removed from counting.

### 2.3.2 The co-occurrence of animals

Based on the list of animals counted using the method described above, we conducted the co-occurrence analyses. To analyse the co-occurrence of the animals, we did not simply count the combination and use the sum value, but we adjusted the value for the following reason. The characters in a folktale can be different in different regions; for example, ‘coyote’ can be substituted as ‘fox’ is different regions. These pair of animals in a substitutional relationship do not indicate that these animals appear in a story together as a co-occurrent pair. Thus, we subtracted the number of all substitutable pairs from the number of all of the combinations in the text.

### 2.3.3 Motifs

As described above, the Thompson Motif Index (TMI) classified motifs into major categories using alphabetical letters (see Figure 3 for each motif description), and we extracted TMI-tagged motifs in the corpus by using Python. In the analysis, we used the major categories represented by single letters (i.e., A to Z). We counted motifs like

“different word count” in NLP. Even if the same motifs appeared more than once in a tale type (e.g. K.1 and K.1), we counted it once; but we counted more than once if the subordinate number was different (e.g. K.1 and K.2.) (See Figure 3).

## 3. Results

### 3.1. Co-occurrence analysis

To investigate the relationships between animals, we calculated the co-occurrence frequency of animals. Note that the co-occurrence frequency here refers to animals appearing at the same time in one tale type, and no duplicate count was made even if it appears more than once in the type (see 2.3). The co-occurrence frequency of animals in the entire corpus is illustrated in Figure 1.

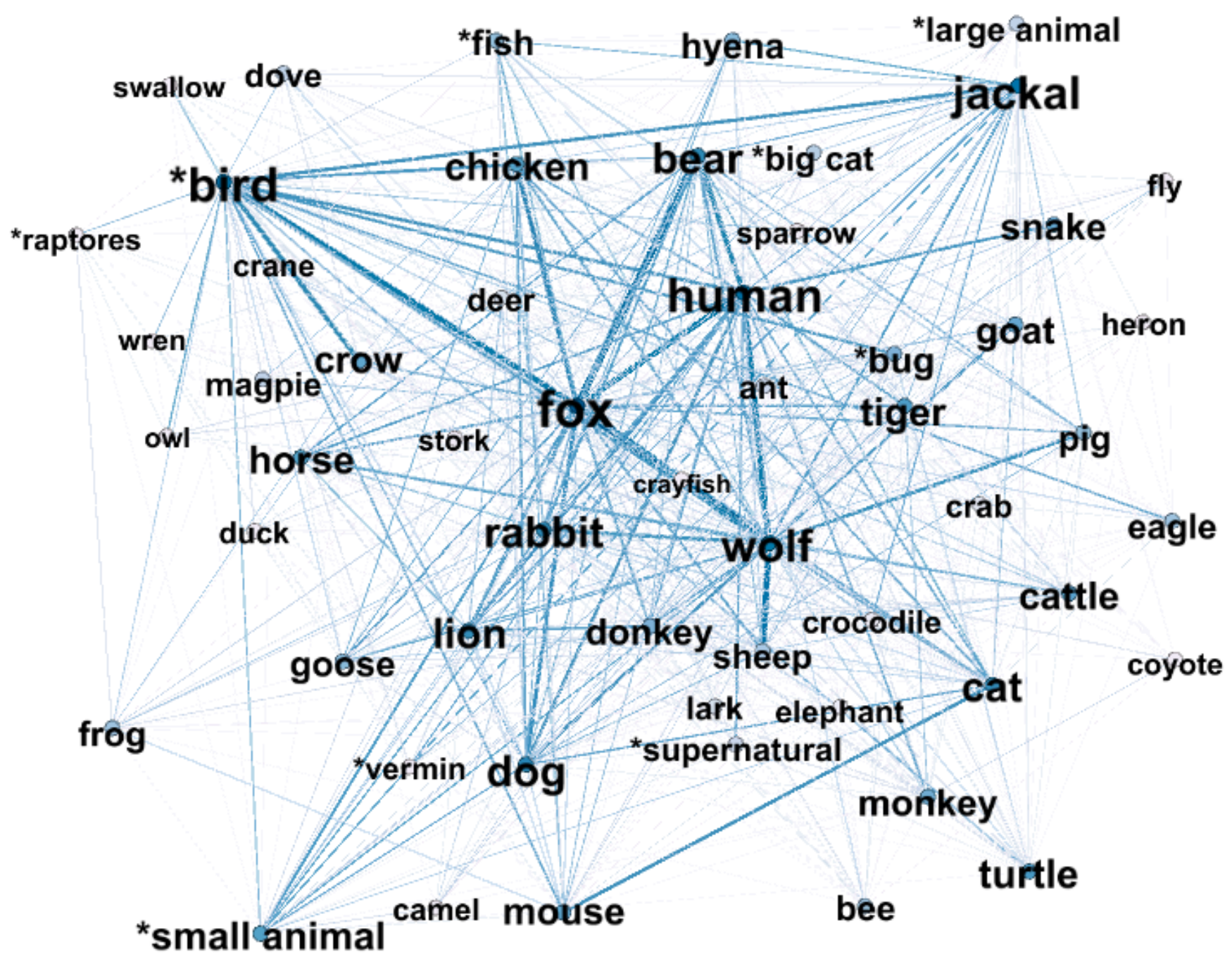

Figure 1. Co-occurrence relationship of the animals. In this figure, the size of each label (e.g., dog or rabbit) represents the sum of the frequency of an animal that appeared co-occurrently. Each link between two labels represents a co-occurrent pair of animals, and the thickness of the links was adjusted by relative frequency. The labels with an asterisk (*) represent integrated categories of low-frequent animals. For example, ‘*bird’ contains words such as ‘swan’ or ‘cuckoo’ that appear at a lower frequency, and it does not include ‘chicken’ or ‘crow’ appears frequently (see 2.3.1. for detail).

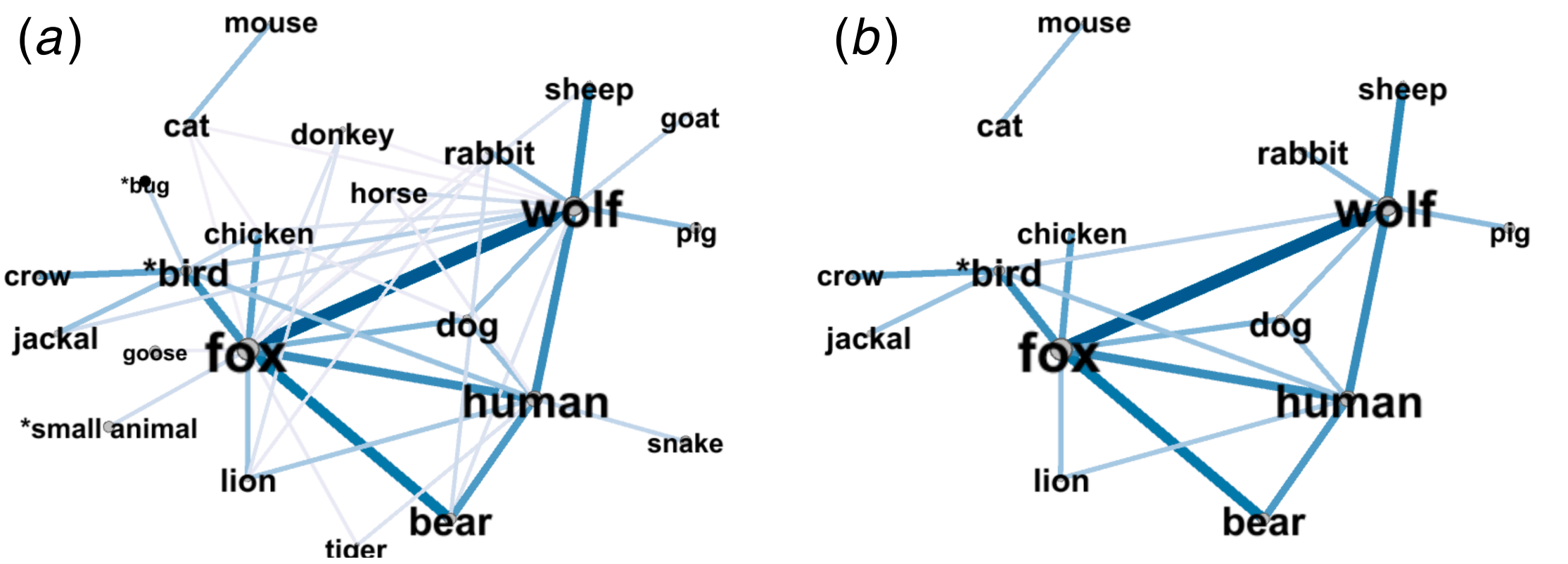

Figure 2. Co-occurrence relationship of animals. (a): The co-occurrence network of animals with a frequency higher than 10. The density of the colours and width of the lines corresponds to the frequency of co-occurrence. (b): The co-occurrence network of animals with a frequency higher than 14. Figure 2 was informative, and so we then extracted the pairs of animals with a high frequency (Figure 2. (a) n >10; Figure 2. (b) n > 14). The figure demonstrated that common adversarial animal relationships, such as cat v. mouse, are often described as predator-prey relationships [37,38]. In addition, foxes were associated with poultry and small animals such as chicken or birds, and wolves were associated with large livestock such as pigs, sheep, and goats. These types of animal attacks have often been reported in livestock protection or food web studies (e.g., [39–41]). Thus, as we discussed in the introduction, pairs of wild animals v. domestic animals appeared in the co-occurrence network, as represented in predator-prey relationships in the real world.

### 3.2. Motif analysis

#### 3.2.1. The frequency of occurrence of the motif

Above, we demonstrated that the pairs of animals, such as domestic v. the wild, appeared in the co-occurrence analysis. Yet, this does not explain the full 'story' of if the pairs of animal characters were in adversarial relationships. Thus, in order to infer the content of the story, we counted the frequency of the motifs (Figure 3).

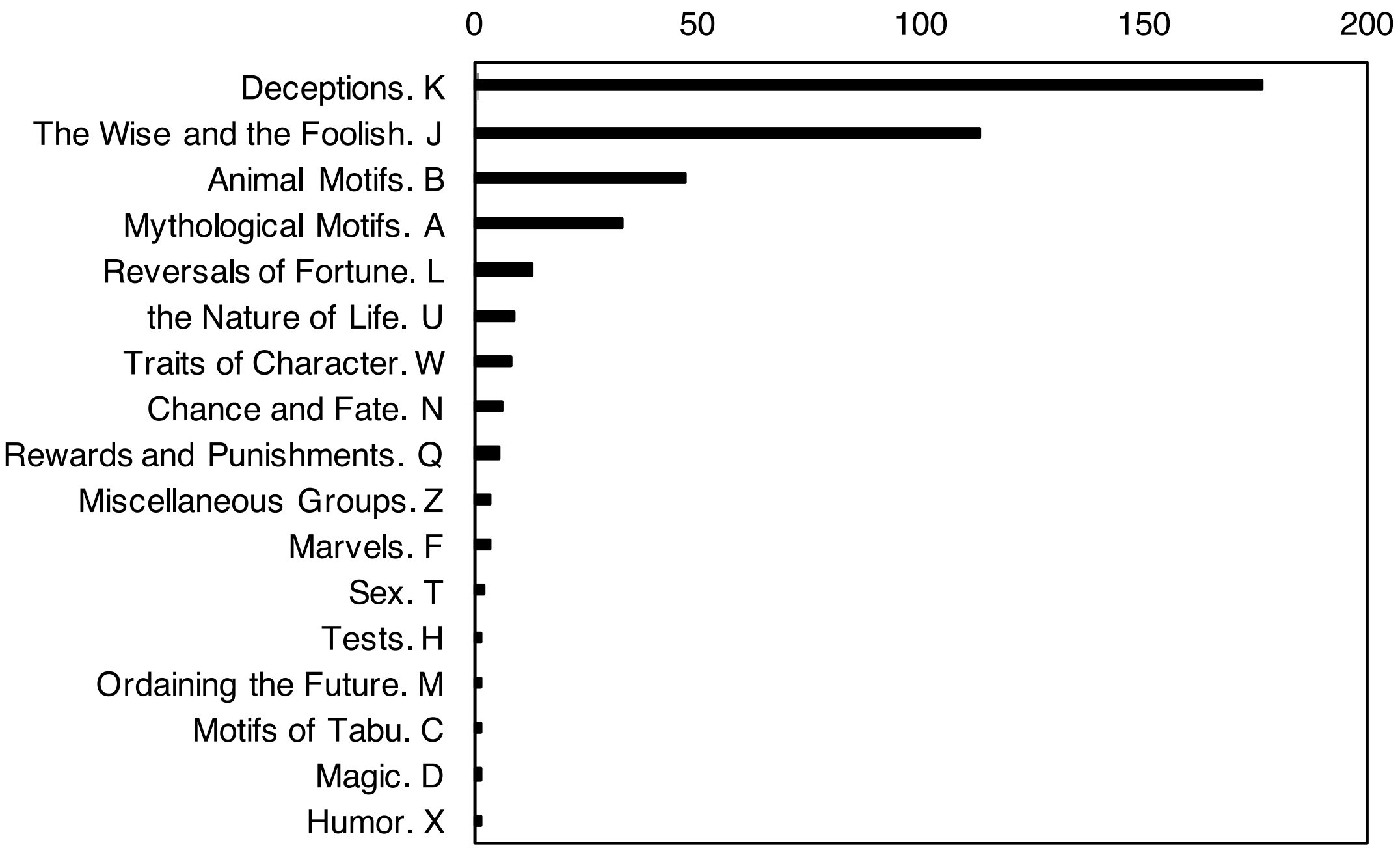

Figure 3. Frequency of the motifs (Thompson Motif Index). We used the original label of motifs used by Thompson [34]

The most frequently appeared motif was K (deceptions). These are motifs where the character deceives another character such as 'Deceptions through shams' (K1700-2099) in ATU123 '*The wolf and the kids*'. The second most common motif was J (the wise and the foolish), which are motifs when wisdom or stupidity leads to a consequence, such as the 'Absurd misunderstanding' (J1750 - J1849) in ATU34A '*The dog drops his meat for the reflection*'. The third most frequent motif was B (the animal motif), which includes motifs such as 'Animals with human traits' (B200 - B299) in ATU222 '*War between birds (insects) and quadrupeds*'. We considered the motif K (deception) as reflective of the adversarial relationships of animals, such as in the animal kingdom when one animal deceives another animal (e.g., through mimicry or strategic deception).

### 3.2.2. The relationship between the classification of folk tales and motifs

In our hypothesis, we considered that folktales contain knowledge about the predator-prey relationship. Thus, the adversarial motif K that implies the predator-prey relationship should appear in accordance with real-world predator-prey relationships. First, we investigated this accordance of the motif and category of folktales, which contain information whether the animal characters were wild or domestic. In the corpus, the tale types were classified into five subcategories according to the type of animals that appear in the story: ‘Wild animals’, ‘wild and domestic animals’, ‘wild animals and humans’, ‘domestic animals’, and ‘others’. In order to investigate whether the distribution of the motifs was different in these subcategories or not, we calculated the relative frequency of the motifs in each subcategory and centred the values by subtracting each motifs’ mean. Figure 4 illustrated that K (deceptions) motifs appeared more frequently in ‘wild animals’ and ‘wild animals and domestic animals’, and less frequently in ‘domestic animals’ and ‘others’. In ‘domestic animals’, J (the wise and the foolish) motifs appeared frequently. In ‘wild animals and humans’, B (animal) motifs appeared frequently.

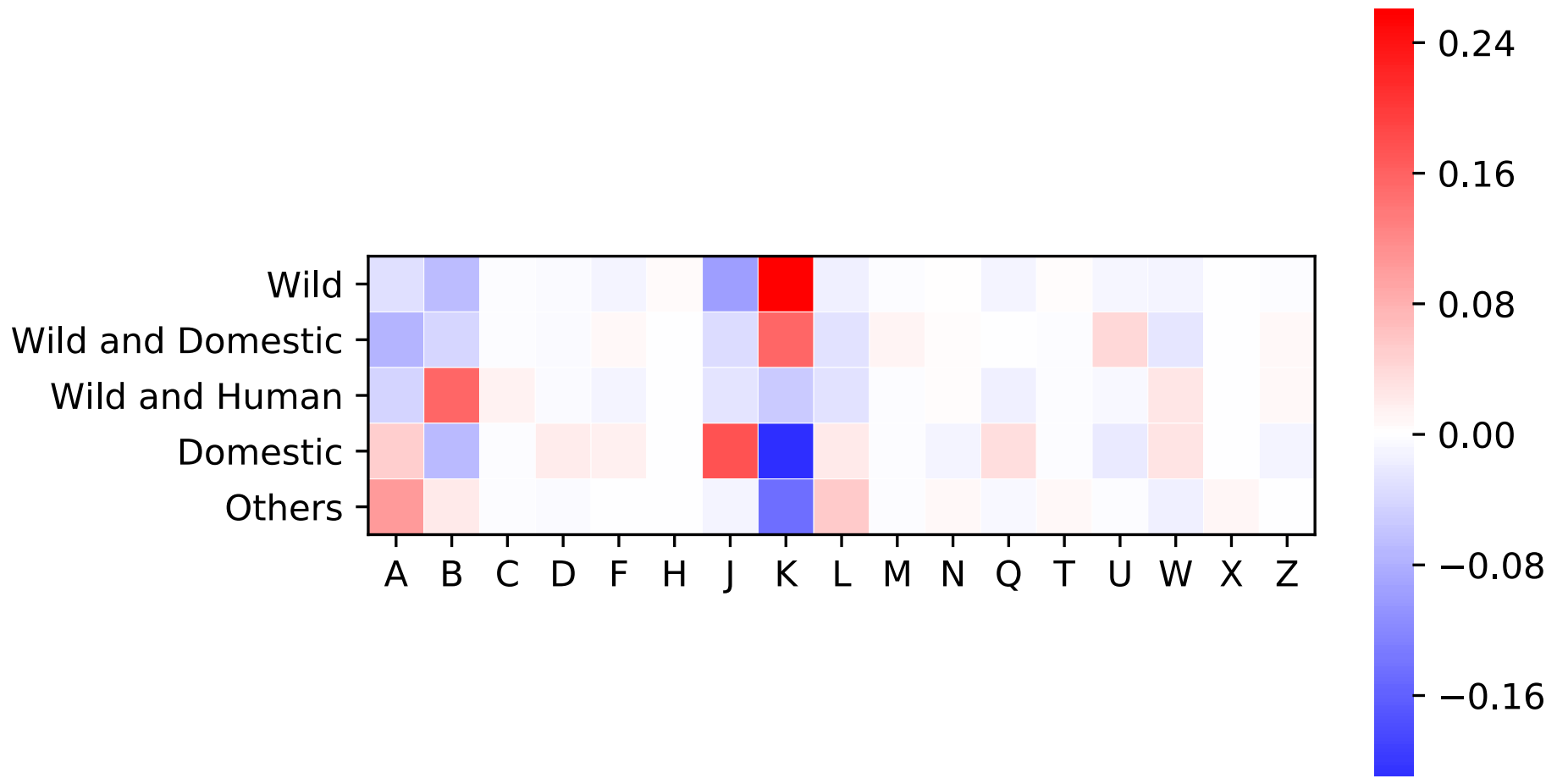

Figure 4. Relative frequency of motifs by classification (after centring). The horizontal label A-Z represents the motifs (see Figure 3 for details), and the vertical axis represents each category. The colour of each label illustrates the relative frequency of the motifs and the relative frequency within each classification were centred by subtracting the average values of each motif.

Here, we focus on motif K, which we considered as a reflection of the predator-prey relationship. Relatively, this motif K appeared in the category of ‘wild’ and ‘wild and domestic’, as most of the pair of animals that appeared in the co-occurrence network were either among wild animals or wild versus domestic animals.

We further verified this point using a principal component analysis (PCA) to examine the variation pattern of the motifs according to the categories. We mapped the variation pattern in two dimensions in Figure 5, and as our objective was to map these into fewer dimensions, we did not interpret the component of these axes. The cumulative contribution rate up to the second principal component was 0.92, and the variance of the data was almost explained by these two axes. In the figure, 'wild animals' and 'wild animals and domestic animals' were placed close to each other, and they were laid in the direction of the same vector K (deceptions). Contrarily, 'domestic animals' was placed in the direction of vector J (the wise and the foolish), and 'wild animals and humans' was placed in the direction of vector B (animal motifs). The result of the PCA analysis was consistent with the relative frequency of the motifs according to the categories.

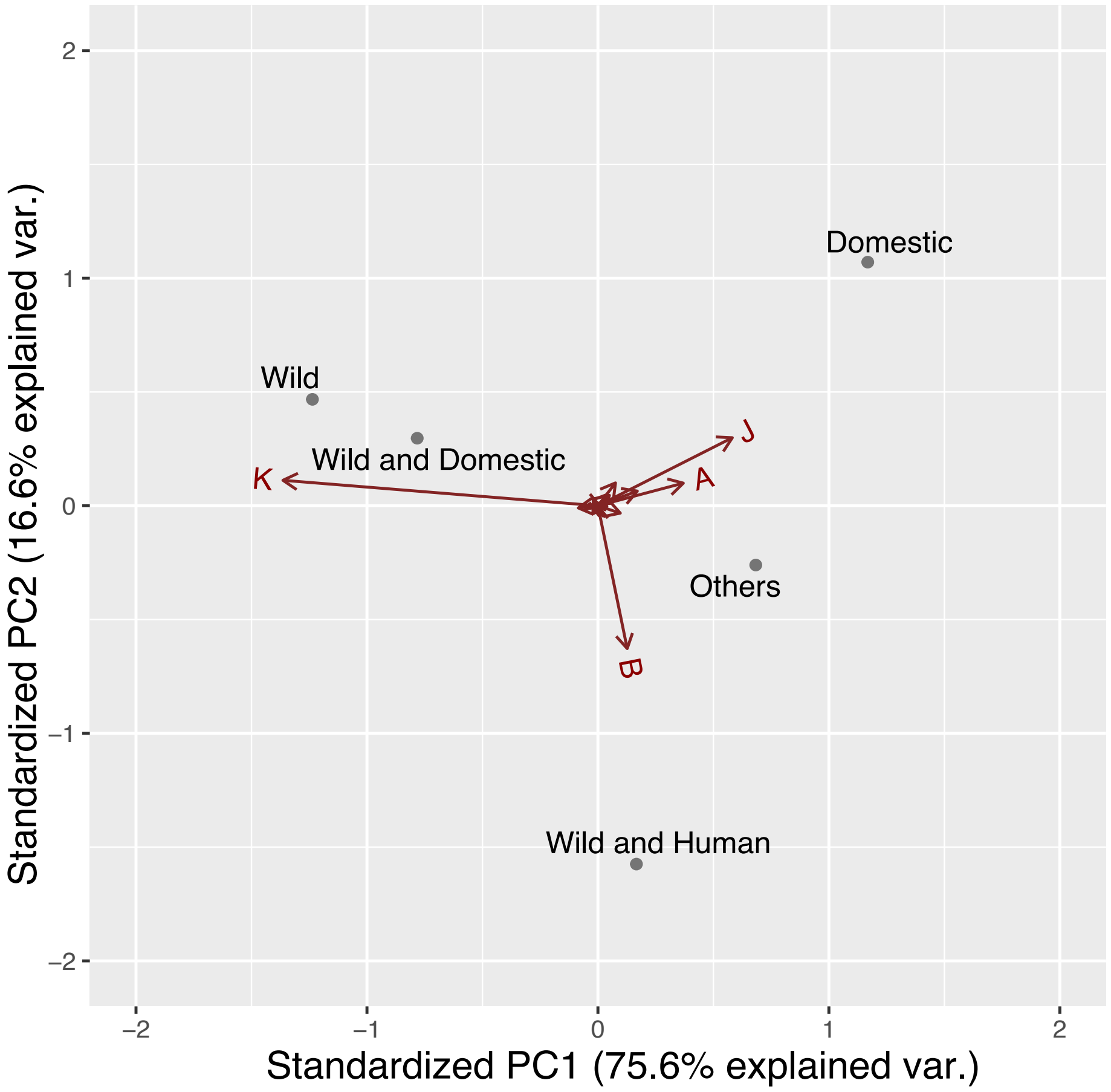

Figure 5. Biplot of motif frequency by category

### 3.2.3. Relationship between individual animals and motifs

Furthermore, instead of the animal category, we analysed the pattern of the individual animals similarly. Thus, we conducted a PCA on the animal occurrence with the motifs (Figure 6). The cumulative contribution rate was 0.87 for the second principal component, and this explained the variance of the data sufficiently. Typical wild animals such as hyenas and jackals, foxes, wolves and their prey chicken, goats, and sheep were placed in a high position of the K (deception) vector, which is consistent with the pairs in the co-occurrence network (see section 3.1.2). However, donkeys, horses, and cantles, two animals that demonstrated the direction of J (wisdom and foolishness) did not have a strong relationship in the co-occurrence network (they appeared in Figure 2. (a) [n > 10] but not in Figure 2. (b) [n > 14]).

The series of analyses of the motifs and categories (relative frequency and PCA) demonstrated that a certain category was associated with specific motifs. Further, the PCA on individual animals demonstrated animals that frequently appeared in pairs were associated with motif K. This suggested that adversarial pairs of animals frequently appeared in the folktales, similar to the natural environment (see also Figure S1).

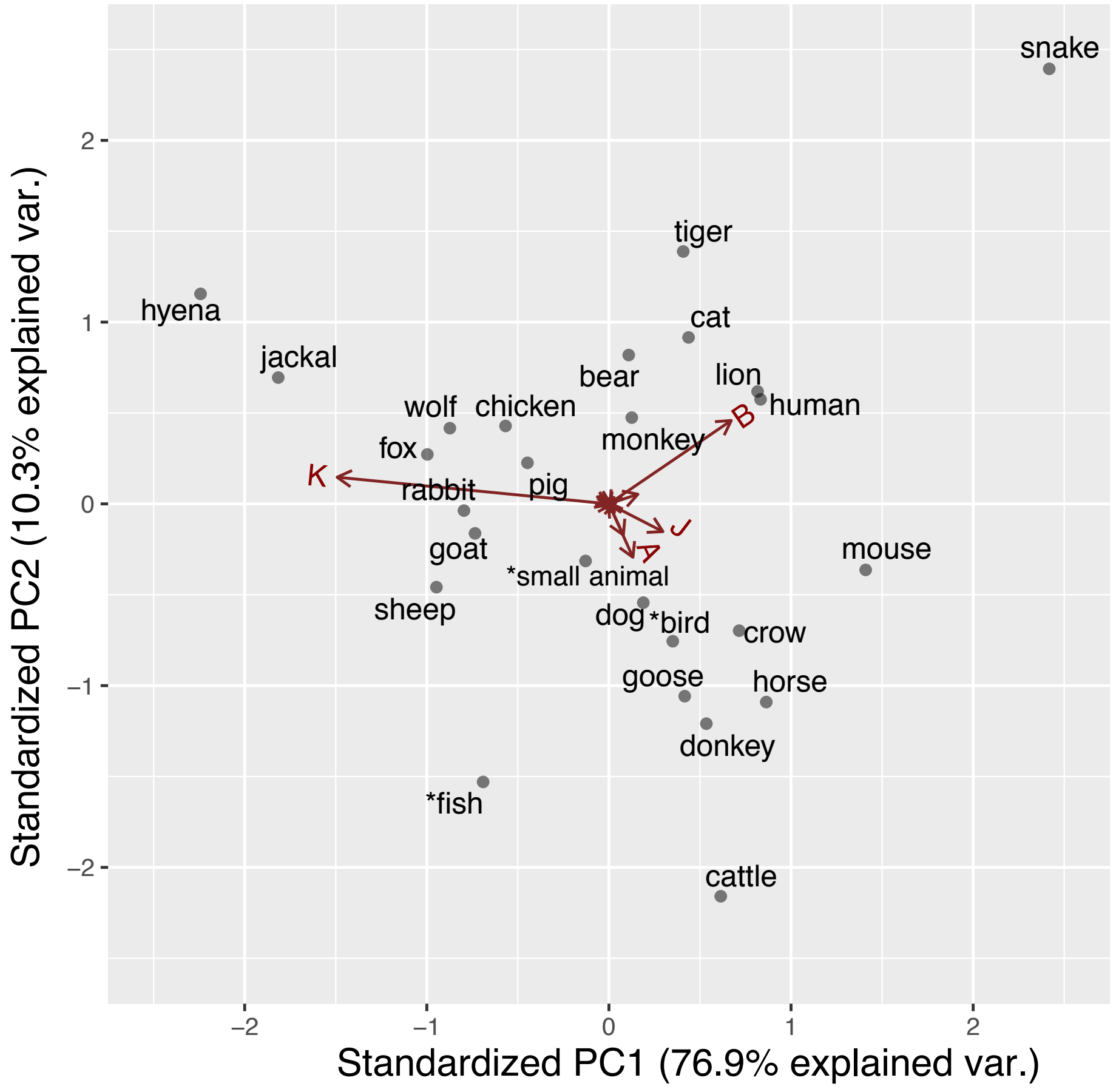

Figure 6. Biplot of the frequency of the animals and motifs (animals where freq. > 30)

## 4. Discussion

### 4.1. Summary of the results

The present study quantitatively analysed folktales in the ATU index by focusing on (1) the co-occurrence network of the animals and the (2) motif analysis of the category and individual animals. The animal pairs that appeared in the co-occurrence network fit well with the real pattern of predator-prey relationships (e.g., fox v. chicken, wolf v, sheep) in the real world. Furthermore, the motif analysis revealed, that the relative frequency of the motifs was different across the folktale categories (i.e., domesticated, wild, wild-domesticated, wild-human, and others). This discrepancy of the relative frequency appeared especially in K ('deception') and J ('the wise and foolish'): the motif K frequently appeared in the category of 'wild' and 'wild and domestic', and the motif J frequently appeared in the category of 'domestic'. The result was consistent with the principal component analysis of the motif by categories and this further confirmed the results. Second, we conducted a principal component analysis of the motifs at the level of individual animals, and this was consistent with previous categories. The results demonstrated that the predator-prey pairs of animals that appeared in the co-occurrence network (e.g., fox v. chicken, wolf v. sheep) was in the direction of the motif of adversarial relationships (K: 'deception). This suggests that the adversarial pairs of animals in predator-prey relationships frequently appeared in folktales.

**4.2. Further discussion of motifs**

**4.2.1 Difference between K and J**

Above, we considered that motif K (deception) was adversarial but not for J (the wise and the foolish). The two motifs seem to be similar in the sense that one character was fooled or behaved foolishly; however, they are different in the relationality of the characters. While deception (motif K) always requires at least two characters, the deceiving and the deceived, foolish actions (motif J) can be achieved alone. Here, we illustrate this difference using specific examples from folktales and explain how motif K represents an adversarial relationship, but motif J does not.

First, we illustrate the dyadic advisory relationship in K. For example, in ATU 127A *'The wolf induces the goat to come down from a cliff and devours it'*, a wild animal (the wolf) deceived a domestic animal (the goat). Further, in ATU 126 '*The sheep chases the wolf* (the sheep pretended to eat the wolf)', the relationship is reversed, and a domestic animal (the sheep) deceived a wild animal (the wolf). This type of adversarial predator-prey relationship is common in the wild, and the opposite part of the relationship can be domestic animals. However, the story describing the relationships among domestic animals should not be applied. Instead of motif K, stories among domestic animals are likely tagged with motif J. For example, in ATU 211 '*The two donkeys and their loads*', two donkeys appear, which are domestic animals, but they are not in an adversarial relationship. In the story, two donkeys are carrying a load for each other, and one donkey goes into some water. As it was carrying salt, the salt dissolved and its load became lighter. The other donkey witnessed the event and attempted to mimic the actions and lighten its load as well. However, this donkey was carrying flour and so their load became heavier instead. Thus, although the two domestic animals appeared in the story, there was no conflict as with predator-prey relationships.

As we demonstrated above, the relation of animals in J and K is different. In motif K, the wild and domestic animals were in an adversarial relationship, but not for the domestic animals in motif J. This is because domestic animals are rarely in adversary relationships in the real world, and folktales of domestic animals may represent this pattern. However, wild animals have adversary relationships, and 'deception' can be observed in the animal kingdom among predator-prey relationships. While in reality, wolves do not use language to hunt goats, nor do sheep pretend to eat wolves, various deceptions can be observed in the animal kingdom, such as mimicry [42] and distractive display (e.g., distracting the attention of predators from younger individuals) [43].

#### 4.2.2 Other motifs

So far, we have not discussed other motifs, such as motif A ('mythical motif') or B ('animal motif'). The relative frequency of A and B were also different according to the categories. Motif A ('mythical motif') frequently appeared in the category of 'domestic' animals. Most of the subordinate motif were items in A 2200-2259, which was about 'Various causes of animal characteristics'. For example, ATU 200B '*Why dogs sniff at one another?*' explained the cause of animal habits. Although these explanations are different from the evolutionary cause of certain animal behaviours, the origin of the actions was still interesting. While we could not explain why this motif appeared in domestic animals specifically, one possibility is that as we need to manage domestic animals in everyday life, retaining the knowledge of the behavioural characteristics of domestic animals may be more important than retaining knowledge about wild animals.

For motif B, unlike motif A, we found various subordinate motifs such as B200-299 'animals with human traits' (e.g., ATU235*: *the animals quarrel*), B500-599 'services of helpful animals' (e.g., ATU201D *: *dogs bark at the thieves*), B700-799 'fanciful traits of animals' (e.g., ATU184: *monkeys always copy man*), and B800-899 miscellaneous animal motifs. As the large category 'animal motif' was varied for the subordinate motifs and certain subordinate categories did not consistently appear, we could not interpret the relative frequency of motif B.

### 4.3. Adaptive implication of knowledge transmission through stories

Finally, we would like to discuss the evolutionary implication of the present study. Researchers have argued that folktales have an adaptive function to promote cooperative behaviours and contain the knowledge of the natural environment, which is beneficial for foraging or agriculture. The present research revealed the folk-zoological knowledge embedded in folktales, including the knowledge essential for survival. However, why are folktales used to transmit this knowledge?

In modern society, we use various information sources: such as books, TV programmes, or the Internet. However, when we did not have either printing systems or systematic education systems, animal folktales may have played the role of obtaining and sharing animal knowledge. From an early age, children learn about animals, and although they rarely observe real animals (such as wolves or lions), children are aware that wolves are dangerous carnivores (e.g., from folk tales such as 'Little Red Riding Hood'), and this is because of the repetitive motif that appears in folktales [3,23,44].

The reason that folktales have been used to transmit information is not only because we did not have any other options, but folktales have beneficial features for transmitting knowledge. Storytelling is more than simply sharing information in a declarative format such as 'Animal *X* is dangerous' for two reasons [10]. First, folktales often contain ritualised styles, such as repetition or rhythm [6,10,24,25]. These repetitive or rhythmic styles are a feature of infant-directed speech, known as 'motherese' [45], which has been demonstrated as a method for attracting the attention of infants and children and it helps their learning [46]. Second, folktales often contain attractive cognitive motifs such as counter-intuitive or humorous events/characters. Animal folktales often include a combination of these cognitively attractive elements to increase memorability, and across cultures, animals with counterintuitive motifs are found in higher frequencies [26].

However, one of the limitations of the current research is that folktales do not mirror the real world perfectly. As we discussed above, folktales contain counter-intuitive elements such as talkative animals. Furthermore, some animals have been symbolised or caricaturized through cultural transmission [23]. For example, the 'fox' or 'jackal' are often known as 'tricksters' with deceptive behaviours [47]. As experimental studies have demonstrated, certain peculiar traits tend to be exaggerated during the transmission [48]. For example, the deceptive behaviours of foxes may simply reflect real-world behaviours toward livestock or other prey. However, this association between 'fox' and 'deception' may be reinforced through repetitive transmission through generations, and so 'fox' becomes a symbol for deception without context, such as 'crafty as a fox'. In this study, we have not analysed these culturally symbolised aspects of folktales, which other anthropologists have discussed. Thus, further analyses in this direction are encouraged.

**4.4. Perspective**

In the present study, we analysed the content of the ATU using natural language processing and statistical analyses. Previous studies in cultural evolution used the ATU to infer geometric patterns, population size, or cultural roots. However, some researchers have argued that the ATU is biased as most stories were collected in Europe [49]. We did not conduct analyses using regional or quantity of information but used abstract information such as motif and category. Thus, we consider that this influence is weaker. While we used the ATU in the initial study because the ATU is an international dataset and its abstracts are tagged with motifs and categorised, further studies are necessary for regional-specific datasets of folktales or other datasets.

Natural language processing is useful in the study of cultural evolution. Morin and Acerbi (2017) [50] combined both theory of cultural evolution and the natural language processing method to understand the shifts in emotional expression in the nineteenth century. In folkloristics, quantitative approaches using NLP or other statistical methods have been introduced as 'computational folkloristics' [32], and they form a bridge between qualitative data and quantitative methods. This study also illustrated the effectiveness of the quantitative approach to qualitative data.

Recently, quantitative analyses of large data have advanced our understanding of the humanities. Historians and anthropologists have analysed the accumulation of archaeological and historical datasets to test theories in social evolutions (e.g., [51]). Similarly, our research tested an anthropology theory using quantitative analyses on quantitative datasets that have been accumulated by folklorists. Rapid technology advances have enabled us to answer new questions by using a rich accumulation of knowledge in humanities and social science, and interdisciplinary studies could answer potential questions that will develop.

**Acknowledgement:**

We thank Dr Christopher Kavanagh for commenting on an earlier version of this article. We would like to thank Editage (www.editage.com) for English language editing. The research was funded by the Toshiba International Foundation.

**Author Contribution:**

The initial research was designed and conducted by Y.N. and K.S., who programmed and analysed the data. Y.N. and K.S. wrote the paper

**Data Availability:**

All the programming codes and the datasets we processed are available at https://github.com/satocos135/animalfolktale-analysis (The original text of Uther 2004 is not uploaded due to the copyright).

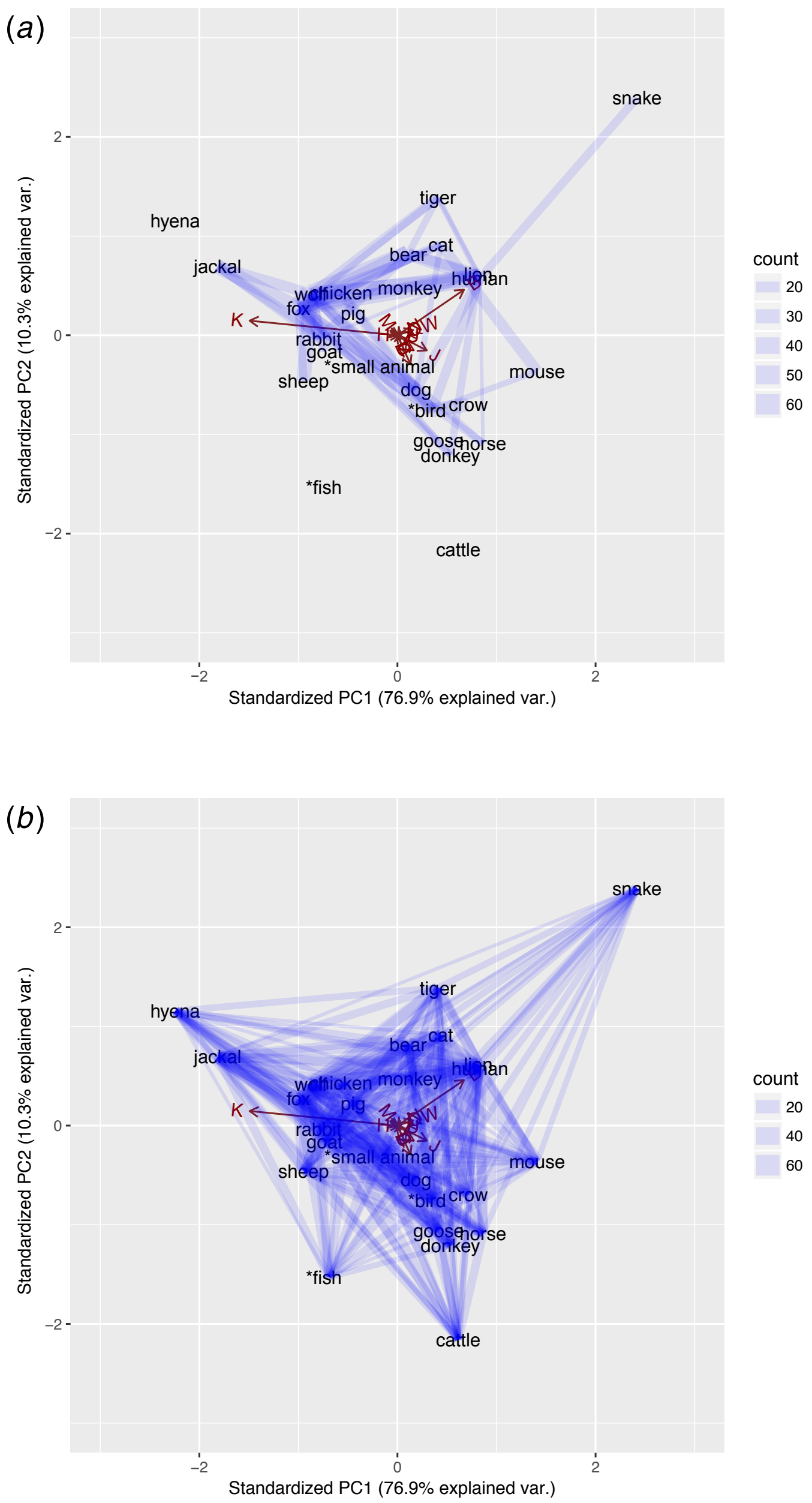

Supplementary Figure S1. Co-occurrence network over biplot of Frequency of Animals and Motifs. (a) Co-occurrent network with frequency more than 10. (b) The entire co-occurrent network (i.e. n > 0). The figures show the three hubs for co-occurrence network. The first hub is around fox, wolf, pig, jackal and sheep, which seems to show domesticated animals and its predators. The second hub is mainly around domestic animals, goose, horse, donkey, dog. The third hub is not as clear as other two clusters but it seems to be centered around human.